\documentclass[review]{elsarticle}

\usepackage{lineno,hyperref}

\journal{Neurocomputing}
\usepackage{graphicx}
\usepackage{algorithm}
\usepackage{algpseudocode}
\usepackage{mathptmx}
\usepackage{multirow}
\usepackage[normalem]{ulem}
\usepackage{vcell}
\usepackage{booktabs}
\usepackage{xcolor}
\makeatletter
\newcommand{\printfnsymbol}[1]{%
  \textsuperscript{\@fnsymbol{#1}}%
}

\begin{document}

\begin{frontmatter}


\title{A Heuristic-driven Uncertainty based Ensemble Framework for Fake News Detection in Tweets and News Articles}


\author[mymainaddress]{Sourya Dipta Das}
\ead[url]{dipta.juetce@gmail.com}

\author[mymainaddress]{Ayan Basak}
\ead[url]{ayan.basak@razorthink.com}

\author[mysecondaryaddress]{Saikat Dutta}
\ead[url]{saikat.dutta779@gmail.com}


\address[mymainaddress]{Razorthink Inc, USA}
\address[mysecondaryaddress]{IIT Madras, India}

\begin{abstract}
The significance of social media has increased manifold in the past few decades as it helps people from even the most remote corners of the world to stay connected. With the advent of technology, digital media has become more relevant and widely used than ever before and along with this, there has been a resurgence in the circulation of fake news and tweets that demand immediate attention. In this paper, we describe a novel Fake News Detection system that automatically identifies whether a news item is “real” or “fake”, as an extension of our work in the  CONSTRAINT COVID-19 Fake News Detection in English challenge. We have used an ensemble model consisting of pre-trained models followed by a statistical feature fusion network , along with a novel heuristic algorithm by incorporating various attributes present in news items or tweets like source, username handles, URL domains and authors as statistical feature. Our proposed framework have also quantified reliable predictive uncertainty along with proper class output confidence level for the classification task.
We have evaluated our results on the COVID-19 Fake News dataset and FakeNewsNet dataset to show the effectiveness of the proposed algorithm on detecting fake news in short news content as well as in news articles. 
We obtained a best F1-score of \textbf{0.9892} on the COVID-19 dataset, and an F1-score of \textbf{0.9156} on the FakeNewsNet dataset. 

\end{abstract}



\end{frontmatter}


%
\section{Introduction}
Fake news represents the press that is used to spread false information and hoaxes through conventional platforms as well as online ones, mainly social media. There has been an increasing interest in fake news on social media due to the political climate prevailing in the modern world~\cite{Tucker_et_al.,Calvillo_et_al,Monti_et_al}, as well as several other factors. Detecting misinformation on social media is as important as it is technically challenging. The difficulty is partly due to the fact that even humans cannot accurately distinguish false from true news, mainly because it involves tedious evidence collection as well as careful fact checking. With the advent of technology and ever-increasing propagation of fake articles in social media, it has become really important to come up with automated frameworks for fake news identification.

In this paper, we describe our system which performs a binary classification on news items from social media and classifies it into “real” or “fake”. We have used transfer learning in our approach as it has proven to be extremely effective in text classification tasks, with a reduced training time as we do not need to train each model from scratch. The primary steps for our approach initially include text preprocessing, tokenization, model prediction, and ensemble creation using a soft voting schema. After completion of the competition, we have drastically improved our fake news detection framework with a Statistical feature fusion network (SFFN) with uncertainty estimation, and followed by a heuristic post-processing technique where both network takes into account the effect of important aspects of news items like username handles, URL domains, news source, news author, etc as statistical features. This approach has allowed us to produce much superior results when compared to other models in their respective datasets. We have also  provided performance analysis of predictive uncertainty quality with proper metrics and showed improvement in overall performance, robustness of the SFFN with ablation study. We have also additionally performed an ablation study of the various attributes used in our post-processing approach. 

Our algorithm is also applicable to detection of fake news items in long news articles. In this context, we have evaluated the performance of our approach on FakeNewsNet dataset \cite{shu2020fakenewsnet}. Along with the news titles,
we have also utilized the actual news body (document) in this case.
We have used a BERT-inspired longformer~\cite{beltagy2020longformer} network which we trained on news articles for classification tasks. We denote this model as NewsBERT in this paper. NewsBERT is used on the news articles to obtain the prediction vectors, which can be used as additional features for our model. After that, we have implemented the same pipeline consisting of SFFN and heuristic post-processing module to boost our performance in FakeNewsNet Dataset.
Using these additional features and modules, we have observed absolute improvement of 9.56 \% in the overall accuracy and F1-score over current state of the art model in FakeNewsNet Dataset. We have also quantified the model uncertainty in the fake news classification task for both datasets.

\section{Related Work}
\subsection{Fake News Detection}
Traditional machine learning approaches have been quite successful in fake news identification problem. Reis et al.~\cite{Reis_et_al} has used feature engineering to generate hand-crafted features like syntactic features, semantic features etc. 
The problem was then approached as a binary classification problem where these features were fed into conventional Machine Learning classifiers like K-Nearest Neighbor (KNN)~\cite{kozma2008k}, Random Forest (RF)~\cite{breiman2001random}, Naive Bayes~\cite{rish2001empirical}, Support Vector Machine (SVM)~\cite{hearst1998support} and XGBOOST (XGB)~\cite{chen2016xgboost}, where RF and XGB yielded results that were quite favourable.
Shu et al.~\cite{Shu_et_al} have proposed a novel framework TriFN, which provides a principled way to model tri-relationship among publishers, news pieces, and users simultaneously. This framework significantly outperformed the baseline Machine Learning models as well as erstwhile state-of-the-art frameworks on early version of FakeNewsNet dataset~\cite{shu2017exploiting}. With the advent of deep learning, there has been a significant revolution in the field of text classification, and thereby in fake news detection. Karimi et al.~\cite{Karimi_et_al} has proposed a Multi-Source Multi-class Fake News Detection framework that can do automatic feature extraction using Convolution Neural Network (CNN) based models and combine these features coming from multiple sources using an attention mechanism, which has produced much better results than previous approaches that involved hand-crafted features. Zhang et al.~\cite{Zhang_et_al} introduced a new diffusive unit model, namely Gated Diffusive Unit (GDU), that has been used to build a deep diffusive network model to learn the representations of news articles, creators and subjects simultaneously. 
Ruchansky et al.~\cite{Ruchansky_et_al} has proposed a novel Capture-Score-Integrate (CSI) framework that uses an Long Short-term Memory (LSTM) network to capture the temporal spacing of user activity and a doc2vec~\cite{doc2vec} representation of a tweet, along with a neural network based user scoring module to classify the tweet as real or fake. It emphasizes the value of incorporating all three powerful characteristics in the detection of fake news: the tweet content, user source, and article response.
Monti et al.~\cite{Monti_et_al} has shown that social network structure and propagation are important features for fake news detection by implementing  a geometric deep learning framework using Graph Convolutional Networks.	Julio et al.~\cite{reis2019supervised} have used a supervised approach for fake news classification using hand-crafted features like linguistic, lexical, psycholinguistic and semantic features, as well as news source and environmental features. They have applied traditional machine learning models on this data like KNN, Naive Bayes, Random Forest, SVM and XGBOOST, out of which Random Forest and XGBOOST have achieved the best results.
Zellers et al.~\cite{zellers2019defending} have introduced a novel fake news generation model, GROVER, that possesses a GPT-like architecture. It has the capability to generate very realistic fake news items in a controlled manner, including various associated meta information like title, news source, publication date, author list, etc. GROVER also outperforms other deep-pretrained models while discriminating between real and fake news articles, hence, it is a powerful model for fake news generation and detection.
Bang et al.~\cite{bang2021model} have tried to develop a robust model for fake news detection that can generalize across different test sets. They have shown their results by performing experiments on two different test sets - FakeNews-19 and Tweets-19. In one approach, they have fine-tuned transformer based language models using robust loss functions, that did not help to improve the F1-score on the FakeNews-19 dataset by much as compared to the traditional cross-entropy loss; however, it showed better generalization on the Tweets-19 dataset. They have also performed an influence-based data cleansing which has improved model robustness and adaptability.
Shu et al.~\cite{shu2019defend} has proposed an automated Fake News Detection framework, dEFEND, that uses a deep hierarchical co-attention network which takes into account the news items and user comments, and provides a classification output along with viable explanations. 



Felber~\cite{felber2021constraint} has analyzed the performance of some classical Machine Learning models using several linguistic features such as n-gram, readability, emotional tone and punctuation along with various preprocessing techniques like stop word removal, stemming/lemmatization, link removal.
Shushkevich et al.~\cite{shushkevich2021tudublin} has used an ensemble technique consisting of Bidirectional LSTM (Bi-LSTM), SVM, Logistic Regression, Naive Bayes. Their combination of Logistic Regression and Naive Bayes models has produced results that are within $5\%$ of state-of-the art results on the given dataset.
Sharif et al.~\cite{sharif2021combating} have tried out various techniques like SVM, CNN, Bi-LSTM, and CNN+BiLSTM with tf-idf and Word2Vec embedding techniques, where  SVM with tf-idf features has produced the best results.
Gautam et al~\cite{gautam2021fake}. has proposed a solution where they have combined topical distributions obtained using Latent Dirichlet Allocation (LDA) and contextualized representations obtained using XLNet. These features are then passed through a 2-layer Feed Forward Neural Network in order to obtain the final classification output. Li et al.~\cite{li2021exploring} has proposed an ensemble model consisting of various pre-trained models like BERT, RoBERTa, ERNIE, etc. using five-fold five-model cross validation. Their pseudo label algorithm has also been able to improve overall model performance. Bilal et al. \cite{ghanem2021fakeflow} have tried to model the flow of affective information in longer news articles using their framework, FakeFlow. They have evaluated their framework on four real-world datasets and have achieved state-of-the-art results, thereby underscoring the importance of affective information in texts.

\subsection{Language models}
Most of the current state-of-the-art language models are based on Transformer~\cite{Vaswani_et_al} and they have proven to be highly effective in text classification problems. They provide superior results when compared to previous state-of-the-art approaches using techniques like Bi-directional LSTM, Gated Recurrent Unit (GRU) based models etc. Hence, we discuss few state of the art transformer based language models in this section. The introduction of the BERT~\cite{Devlin_et_al} architecture has transformed the capability of transfer learning in Natural Language Processing. It has been able to achieve state-of-the art results on downstream tasks like text classification. RoBERTa~\cite{Liu_et_al} is an improved version of the BERT model. It is derived from BERT’s language-masking strategy, modifying its key hyperparameters, including removing BERT’s next-sentence pre-training objective, and training with much larger mini-batches and learning rates, leading to improved performance on downstream tasks. XLNet~\cite{Yang_et_al} is a generalized auto-regressive language method. It calculates the joint probability of a sequence of tokens based on the transformer architecture having recurrence. Its training objective is to calculate the probability of a word token conditioned on all permutations of word tokens in a sentence, hence capturing a bidirectional context. XLM-RoBERTa~\cite{Conneau_et_al} is a transformer~\cite{Vaswani_et_al} based language model relying on Masked Language Model Objective. DeBERTa~\cite{He_et_al} provides an improvement over the BERT and RoBERTa models using two novel techniques; first, the disentangled attention mechanism, where each word is represented using two vectors that encode its content and position, respectively, and the attention weights among words are computed using disentangled matrices on their contents and relative positions, and second, the output softmax layer is replaced by an enhanced mask decoder to predict the masked tokens pre-training the model. ELECTRA~\cite{Clark_et_al} is used for self-supervised language representation learning. It can be used to pre-train transformer networks using very low compute, and is trained to distinguish ``real" input tokens vs ``fake" input tokens, such as tokens produced by artificial neural networks. ERNIE 2.0~\cite{Sun_et_al.} is a continual pre-training framework to continuously gain improvement on knowledge integration through multi-task learning, enabling it to learn various lexical, syntactic and semantic information through massive data much better.


\subsection{Uncertainty}
Model uncertainty is a very important concept that is related to the model parameters. In order to capture model uncertainty, a prior distribution needs to be assigned over each weight in a neural network. Gal et al.~\cite{gal2016dropout} has developed a new theoretical framework casting dropout training in deep neural networks (NNs) as approximate Bayesian inference in deep Gaussian processes. They have shown that a neural network with arbitrary depth or non-linearities can be analogous to a probabilistic deep Gaussian process when dropout is applied before every weight layer. This theory presents tools to model uncertainty with dropout NNs, and shows a considerable improvement in predictive log-likelihood and Root Mean Squared Error (RMSE) compared to existing state-of-the-art methods.
Lakshminarayanan et al.~\cite{lakshminarayanan2016simple} has proposed a novel approach to estimate the predictive uncertainty using ensembles of deep neural networks. This approach produced superior results when compared to traditional Bayesion Neural Networks, with an added advantage of being readily parallelizable and requiring less hyperparameter tuning. It also takes into account the data uncertainty as it produces higher uncertainty values for out-of-distribution examples.
In the Fake News detection task, uncertainty estimation is a very important aspect since it improves the reliability and safety of the system. It gives us an estimate of how far we can trust a system, and thus increases the interpretability of a system's output. In the case of Fake News detection, it is extremely important to have a system that is both robust and reliable. If visibly benign texts are constantly flagged as fake, it leads to a reduction in credibility of the system. Similarly, if the system fails to identify a lot of fake news items, the scenario becomes dangerous. Hence, uncertainty estimation can provide the user some idea about the fault tolerance level.

\section{Dataset Description}
We have used two datasets to train and evaluate our approach that have the necessary attributes that we require to extract statistical features.

\subsection{COVID-19 Fake News } The dataset~\cite{Patwa_et_al} for CONSTRAINT COVID-19 Fake News Detection in English challenge was provided by the organizers on the competition website\footnote{https://competitions.codalab.org/competitions/26655}. It consists of data that have been collected from various social media and fact checking websites, and the veracity of each post has been verified manually. The “real” news items were collected from verified sources which give useful information about COVID-19, while the “fake” ones were collected from tweets, posts and articles which make speculations about COVID-19 that are verified to be false. Fake allegations have been gathered from a number of fact-checking websites like Politifact, Boomlive, NewsChecker and from other tools like IFCN chatbot and Google fact-check-explorer. Verified Twitter handles like World Health Organization (WHO), Centers for Disease Control and Prevention (CDC), Covid India Seva, Indian Coun- cil of Medical Research (ICMR), etc, are also used to gather real news from Twitter. The original dataset contains 10,700 social media news items, the vocabulary size (i.e., unique words) of which is 37,505 with 5141 words in common to both fake and real news. It is class-wise balanced with 52.34\% of the samples consisting of real news, and 47.66\% of fake samples. These are 880 unique username handle and 210 unique URL domains in the data.

\subsection{FakeNewsNet } We have also evaluated the performance of our fake news detection system on the FakeNewsNet dataset~\cite{shu2020fakenewsnet}, which consists of two datasets with news content, social context, and spatiotemporal information: PolitiFact and GossipCop. In PolitiFact, the political news items are reviewed by journalists and domain experts, who review and provide fact-checking evaluation results to claim news articles as fake or real. 
GossipCop is a website for fact-checking entertainment stories aggregated from various media outlets. GossipCop provides rating scores on the scale of 0 to 10 to classify a news story as the degree from fake to real. Most news items on GossipCop have a rating less than 5, which aligns with its purpose to showcase more fake stories. In order to collect real entertainment news items, the E! Online website~\footnote{https://www.eonline.com/ap} is crawled. It is a well known trusted media website for publishing entertainment news items. The articles from E! Online are considered as real news articles, while the ones from GossipCop are considered fake.
The original dataset consists of 16817 real news items and 5323 fake news items from GossipCop, and 624 real news items and 432 fake news items from PolitiFact. However, we believe that Twitter’s policy to remove certain fake news items from time to time, has prevented us from obtaining the entire dataset. We could able to crawl 15151 real news items and 5323 fake news items from GossipCop, and 610 real news items and 401 fake news items from PolitiFact. Total number of unique news website/news source avalible in this dataset is 2244 and total number of authors avalible in this dataset is 4616. Number of unique keywords for news articles used are 6882. We have done a 80-10-10 split of the data into training, validation and test sets.

\section{Methodology}
Our goal in this paper is to design a common fake news classification pipeline framework for both tweets and news items. For this method, we have used some easily available meta-data of tweets or news to boost the performance of the framework. We are also providing the uncertainty value along with predictions to make this framework suitable for active learning, as well as solving domain adaptation related problems. We have used an ensemble of Pre-trained deep learning based language model for text classification and have fed the prediction vector from that ensemble model to another Approximate Bayesian Neural Network based feature fusion network along with some statistical features computed from meta-data of those news or tweets. Initial prediction vector from that fusion model is further tuned with a heuristic-based post processing approach to boost the qualitative performance of the model. Our proposed method consists of six main parts: (a) Text Preprocessing, (b) Tokenization, (c) Backbone Model Architectures, (d) Ensemble, (e) Statistical Feature Fusion Network, (f) Predictive Uncertainty Estimation Model, and (g) Heuristic Post Processing. The overall architecture of our system is shown in Figure-\ref{initial_process_diag_fig}. A more detailed description is provided in the following subsections:

\subsection{Text Preprocessing}
Some social media items, like tweets, are mostly written in colloquial language. Also, they contain various other information like usernames, URLs, emojis, etc. We have filtered out such attributes from the given data as a basic preprocessing step, before feeding it into the ensemble model. For tweets, We have used the tweet-preprocessor\footnote{pypi.org/project/tweet-preprocessor/} library from Python to filter out such noisy information from tweets. For News articles, we have removed any username, URLs from Instagram, Facebook, Twitter etc.

\subsection{Tokenization}
During tokenization, each sentence is broken down into tokens before being fed into a model. We have used a variety of tokenization approaches\footnote{huggingface.co/docs/tokenizers/python/latest/} depending upon the pre-trained model that we have used, as each model expects tokens to be structured in a particular manner, including the presence of model-specific special tokens. Each model also has its corresponding vocabulary associated with its tokenizer, trained on a large corpus data like GLUE, wikitext-103, CommonCrawl data etc. During training, each model applies the tokenization technique with its corresponding vocabulary on our news data.
We have used a combination of BERT~\cite{Devlin_et_al}, XLNet~\cite{Yang_et_al}, RoBERTa~\cite{Liu_et_al}, XLM-RoBERTa~\cite{Conneau_et_al}, DeBERTa~\cite{He_et_al}, ERNIE 2.0~\cite{Sun_et_al.} and ELECTRA~\cite{Clark_et_al} models and have accordingly used the corresponding tokenizers from the base version of their pre-trained models.

\subsection{Backbone Model Architectures}
We have used a variety of pre-trained language models\footnote{huggingface.co/models} as backbone models for text classification. For each model, an additional fully connected layer is added to its respective encoder sub-network to obtain prediction probabilities for each class- ``real" and ``fake" as a prediction vector. We have used transfer learning in our approach in this problem. Each model has been initialized using pre-trained weights. Thereafter, it fine-tunes the model weights using the tokenized training data. The same tokenizer is used to tokenize the test data and the fine-tuned model checkpoint is used to obtain predictions during inference.

\subsection{Ensemble}
In this method, we use the model prediction vectors obtained from inference on the news titles for the different models to obtain our final classification result, i.e. “real” or “fake”. Our main motivation behind using an ensemble of various fine-tuned pre-trained language models are to utilize knowledge extracted by the respective models from the corresponding dataset in which it is trained on. However, in the case of FakeNewsNet dataset, we obtain an additional prediction vector using NewsBERT on the news body, that is also appended to the existing feature set. All the features used here are obtained from the raw text data only.
To balance an individual model's limitations, an ensemble method can be useful for a collection of similarly well-performing models. We have experimented with two approaches: soft voting and hard voting, that are described in the following figure:

\begin{figure}
\centering
\includegraphics[width=\textwidth]{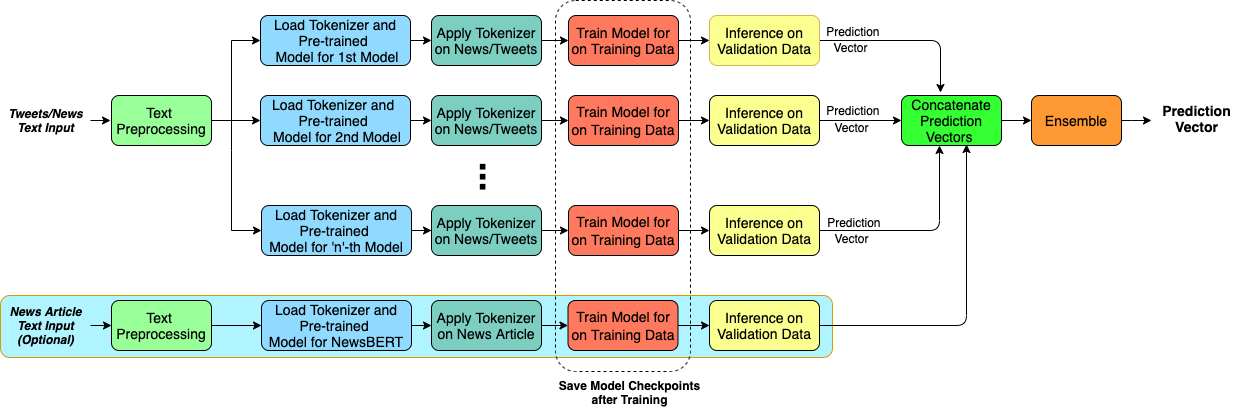}
\caption{Fake News Identification Initial Process Block Diagram} 
\label{initial_process_diag_fig}
\end{figure}
\subsubsection{Soft Voting}
In this approach, we calculate a “soft probability score” for each class by averaging out the prediction probabilities of various models for that class. The class that has a higher average probability value is selected as the final prediction class. 
Probability for ``real" class, $P^r(x)$ and probability for ``fake" class , $P^f(x)$ for a tweet $x$ is given by, 

\begin{equation}\label{equ1}
    P^r(x) = \sum_{i=1}^{n}\frac{P^r_i(x)}{n}
\end{equation}
\begin{equation}\label{equ2}
   P^f(x) = \sum_{i=1}^{n}\frac{P^f_i(x)}{n}
\end{equation}
where $P^r_i(x)$ and $P^f_i(x)$ are ``real" and ``fake" probabilities by the $i$-th model and $n$ is the total number of models. 
\subsubsection{Hard Voting : } 
In this approach, the predicted class label for a news item is the class label that represents the majority of the class labels predicted by each individual model. In other words, the class with the most number of votes is selected as the final prediction class. Votes for ``real" class, $V^r(x)$ and Votes for ``fake" class , $V^f(x)$ for a tweet $x$ is given by,
\begin{equation}\label{equ3}
   V^r(x) = \sum_{i=1}^{n}{I(P^r_i(x) \geq P^f_i(x))}
\end{equation}
\begin{equation}\label{equ4}
   V^f(x) = \sum_{i=1}^{n}{I(P^r_i(x) < P^f_i(x))}
\end{equation}
where the value of $I(a)$ is $1$ if condition $a$ is satisfied and $0$ otherwise. 

\subsection{Statistical Feature Fusion Network}
\label{sec_sffn}
Our basic intuition behind using statistical features is that meta-attributes like username handles, URL domains, news source, news author, etc. are very important aspects of a news item and they can convey reliable information regarding the genuineness of such items. We have tried to incorporate the effect of these attributes along with our original ensemble model predictions. We have calculated probability values corresponding to each of the attributes, for example the probability of an username handle or URL domain indicating a fake news item, and added them to our feature set. We have used information about the frequency of each class for each of these attributes in the training set to compute these probability values. In our experiments, we observed that Soft-voting works better than Hard-voting. Hence our post-processing step takes Soft-voting prediction vectors into account. The steps taken in this approach are described as follows:

\begin{itemize}
    \item   First, we obtain the class-wise probability from the best performing ensemble model. These probability values form two features of our new feature-set.
    
    \item We collect all distinct values of a particular attribute from all the news items in our training data, and calculate how many times the ground truth is “real” or “fake” for this attribute.
    
    \item   We calculate the conditional probability of this particular attribute indicating a real news item, which is represented as follows:
        
        \begin{equation}\label{equ3}
            P^r(x|attribute_{k}) = \frac{n(A)}{n(A) + n(B)}
        \end{equation}
        
    where n(A) = number of “real” news items containing the $attribute_{k}$,  n(B) = number of “fake” news items containing the $attribute_{k}$, and k = 1,2,...,n. In our case, $attribute_{1}$ = "URL domain" in case of COVID-19 Fake News dataset and "news author" in case of FakeNewsNet, and $attribute_{2}$ = "username handle" in case of COVID-19 Fake News dataset and "news source" in case of FakeNewsNet. Similarly, the conditional probability of the particular attribute indicating a fake news item is given by,
        
        \begin{equation}\label{equ4}
            P^f(x|attribute_{k}) = \frac{n(B)}{n(A) + n(B)}
        \end{equation}

        We obtain a probability vector that forms two additional features of our new dataset.

\item  Similarly, we collect all other relevant attributes from all the news items in our training data, and calculate how many times the ground truth is “real” or “fake” for each one. This enables us to compute a two-dimensional prediction vector for each new attribute which can be appended to our current feature set.
This approach enables us to create two types of feature-sets: one using the 10 raw DL model prediction values and the statistical features obtained from various attributes, and the other using the 2 ensembled DL model prediction values and the statistical features. When we use the ensembled prediction features, we obtain superior results. 

        
        
            
\item  In case there are multiple attributes of the same type in a sentence, the final probability vectors are obtained by averaging out the vectors of the individual attribute instances.

\end{itemize}

\subsection{Predictive Uncertainty Estimation Model}
We have designed an approximate Bayesian neural network as a Statistical Feature Fusion Network (SFFN) for uncertainty estimation of fake news classification. We have applied Monte Carlo Dropout (MCDropout)~\cite{gal2016dropout} layer between hidden layers of the feature fusion network for Bayesian interpretation. In the case of Monte Carlo (MC) dropout, the dropout is applied both during training and inference. Hence, the model does not produce the same output each time inference is done on the same data point. Hence, MC dropout enables us to make random predictions that can be interpreted as samples from a probability distribution. As we are using fine tuned Pre-Trained language models for text classification, employing Monte Carlo Dropout (MCDropout) in between the model architecture was not feasible. From this model, we get the prediction vector along with its uncertainty value. 

During inference, we ran multiple forward passes through the trained SFFN model with MCDropout, $f_{SFFN}$ for sample $x$ with different dropout masks.  
For predictive uncertainty estimation, the prediction vector of $N$ inferences with different dropout masks, $d_{0},...,d_{N}$ are accumulated. Here, $f_{SFFN}^{d_{i}}$ represents the model with dropout mask, $d_{i}$.
Hence, for dropout masks ($d_{i}$), we obtain a sample of the possible model outputs, $f_{SFFN}^{d_{0}}(x),....,f_{SFFN}^{d_{N}}(x)$ for that particular sample, $x$.
We get an ensemble prediction by calculating the mean ($\mu_{x}$) and variance ($\sigma^{2}_{x}$) of this sample, which would be the mean of the model's posterior distribution for this sample and an approximation of the model's uncertainty.
\begin{equation}
    v_{p}=\mu_{x}=\frac{1}{N} \sum_{i=0}^{N} f_{SFFN}^{d_{i}}(x)
\end{equation}
\begin{equation}
    c_{u}=\sigma^{2}_{x}=\frac{1}{N}\sum_{i=0}^{N}\left[f_{SFFN}^{d_{i}}(x)-v_{p}\right]^{2}
\end{equation}
Here, $v_{p}$ is the predictive posterior mean and $c_{u}$ is the model uncertainty. 

\subsection{Heuristic Post-Processing}
In this approach, we have augmented our original framework with a heuristic approach that can take into account the effect of the statistical attributes mentioned in Section \ref{sec_sffn}. This approach works well for data having attributes like URL domains, username handles, news source, and the like. Please note, for texts that lack these attributes, we rely only on ensemble model predictions. These attributes allow us to add meaningful features to our current feature set. We obtain new training, validation and test feature-sets obtained using class-wise probability vectors from ensemble model outputs as well as probability values obtained using statistical attributes from the training data. We use a novel heuristic algorithm on this resulting feature set to obtain our final class predictions.
The intuition behind using a heuristic approach taking the statistical features into account is that if a particular feature can by itself be a strong predictor for a particular class, and that particular class is predicted whenever the value of a feature is greater than a particular threshold, a significant number of incorrect predictions obtained using the previous steps can be ``corrected" back.

\begin{figure}
\centering
\includegraphics[width=\textwidth]{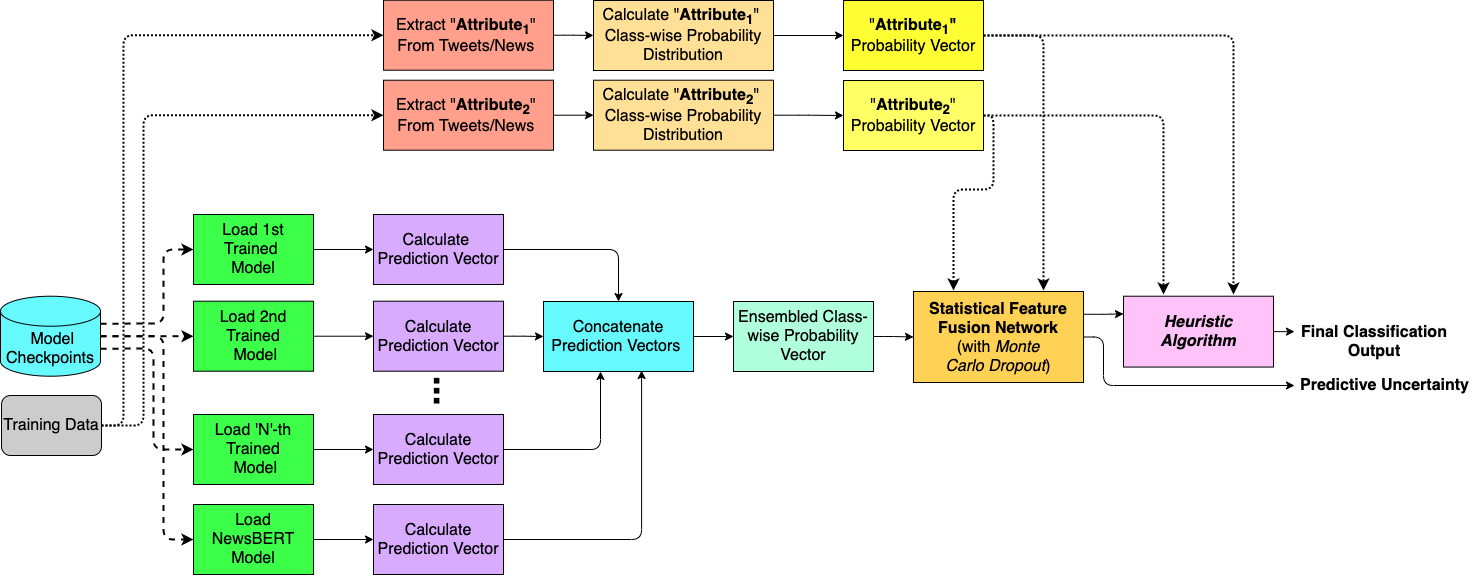}
\caption{Fake News Identification Post Process Block Diagram} 
\label{post_process_diag_fig}
\end{figure}


        
        
            



Table \ref{username domain examples} shows some samples of the conditional probability values of each label class given the URL domain and username handle attributes in COVID-19 Fake News dataset, while Table \ref{author source examples} illustrates similar samples for news source and news author attributes in FakeNewsNet dataset. We have also shown the frequency of those attributes in the training data. The details of the heuristic algorithm is explained in the following pseudocode (Algorithm-\ref{h_algo}). In our experiment, the values of threshold chosen are 0.88 and 0.94 for COVID-19 Fake News Dataset and FakeNewsNet Dataset respectively by using elbow method. The post-processing architecture is shown in Figure-\ref{post_process_diag_fig}.
\begin{algorithm}
	\caption{Heuristic Algorithm}
	\label{h_algo}
	{\textbf{Result: }label ( ``real" or ``fake") }
\begin{algorithmic}[1]
  \If {$P^r(x|attribute_{1}) > threshold $ AND $P^r(x|attribute_{1}) > P^f(x|attribute_{1})$} 
    \State {label = ``real"}
  \ElsIf{$P^f(x|attribute_{1}) > threshold $ AND $P^r(x|attribute_{1}) < P^f(x|attribute_{1})$  }
    \State {label = ``fake"}
  \ElsIf {$P^r(x|attribute_{2}) > threshold $ AND $P^r(x|attribute_{2}) > P^f(x|attribute_{2})$} 
    \State {label = ``real"}
  \ElsIf{$P^f(x|attribute_{2}) > threshold $ AND $P^r(x|attribute_{2}) < P^f(x|attribute_{2})$  }
    \State {label = ``fake"}
  \ElsIf{$P^r(x) > P^f(x)$}
    \State {label = ``real"}
    \Else 
    \State {label = ``fake"}
  \EndIf
\end{algorithmic}
\end{algorithm}

\section{Experiments}
\subsection{Attribute Extraction}
In order to extract statistical features mentioned in Section \ref{sec_sffn}, we have considered the username handles and URL domains from the COVID-19 Fake News dataset and news source and news domain from the FakeNewsNet dataset. Such attributes provide a significant lift in the final classification task since they contribute meaningful information regarding the origin of news items.  

\begin{table}
\centering
\caption{Few Examples on URL Domain-name and Username attribute distribution data }
\label{username domain examples}
\resizebox{\textwidth}{!}{%
\begin{tabular}{|c|c|c|c|l|c|c|c|c|} 
\cline{1-4}\cline{6-9}
\multicolumn{4}{|c|}{Example of URL Domain Name Prob. Dist.} & ~ ~ & \multicolumn{4}{c|}{Example of UserName Prob. Dist.} \\ 
\cline{1-4}\cline{6-9}
\begin{tabular}[c]{@{}c@{}}URL Domain \\Name \end{tabular} & \begin{tabular}[c]{@{}c@{}}$P^r(x|domain)$ \end{tabular} & \begin{tabular}[c]{@{}c@{}}$P^f(x|domain)$ \end{tabular} & \begin{tabular}[c]{@{}c@{}}Frequency \end{tabular} & \multirow{6}{*}{} & UserName & \begin{tabular}[c]{@{}c@{}}$P^r(x|username)$ \end{tabular} & \begin{tabular}[c]{@{}c@{}}$P^f(x|username)$ \end{tabular} & \begin{tabular}[c]{@{}c@{}}Frequency \end{tabular} \\ 
\cline{1-4}\cline{6-9}
news.sky & 1.0 & 0.0 & 274 &  & MoHFW\_NDIA & 0.963 & 0.037 & 162 \\ 
\cline{1-4}\cline{6-9}
medscape.com & 1.0 & 0.0 & 258 &  & DrTedros & 1.0 & 0.0 & 110 \\ 
\cline{1-4}\cline{6-9}
thespoof.com & 0.0 & 1.0 & 253 &  & ICMRDELHI & 0.9903 & 0.0097 & 103 \\ 
\cline{1-4}\cline{6-9}
newsthump.com & 0.0 & 1.0 & 68 &  & PIB\_ndia & 1.0 & 0.0 & 83 \\ 
\cline{1-4}\cline{6-9}
theguardian.com & 0.167 & 0.833 & 6 &  & CDCMMWR & 1.0 & 0.0 & 34 \\
\cline{1-4}\cline{6-9}
\end{tabular}
}
\end{table}

\begin{table}
\centering
\caption{Few Examples on Author and Source attribute distribution data }
\label{author source examples}
\resizebox{\textwidth}{!}{%
\begin{tabular}{|c|c|c|c|l|c|c|c|c|} 
\cline{1-4}\cline{6-9}
\multicolumn{4}{|c|}{Example of Source Prob. Dist.} & ~ ~ & \multicolumn{4}{c|}{Example of Author Prob. Dist.} \\ 
\cline{1-4}\cline{6-9}
\begin{tabular}[c]{@{}c@{}}Source \end{tabular} & \begin{tabular}[c]{@{}c@{}}$P^r(x|source)$ \end{tabular} & \begin{tabular}[c]{@{}c@{}}$P^f(x|source)$ \end{tabular} & \begin{tabular}[c]{@{}c@{}}Frequency \end{tabular} & \multirow{6}{*}{} & Author & \begin{tabular}[c]{@{}c@{}}$P^r(x|author)$ \end{tabular} & \begin{tabular}[c]{@{}c@{}}$P^f(x|author)$ \end{tabular} & \begin{tabular}[c]{@{}c@{}}Frequency \end{tabular} \\ 
\cline{1-4}\cline{6-9}
people.com & 0.8869 & 0.1131 & 1769 &  & Amy Mistretta & 0.1175 & 0.8825 & 434 \\ 
\cline{1-4}\cline{6-9}
www.dailymail.co.uk & 0.8134 & 0.1866 & 943 &  & Lindsay Valdez & 0.1175 & 0.8825 & 4340 \\ 
\cline{1-4}\cline{6-9}
www.usmagazine.com & 0.8063 & 0.1937 & 697 &  & Daisy Maldonado & 0.0681 & 0.9319 & 411 \\ 
\cline{1-4}\cline{6-9}
hollywoodlife.com & 0.8677 & 0.1323 & 446 &  & Dailymail.Com Reporter & 0.8889 & 0.1111 & 243 \\ 
\cline{1-4}\cline{6-9}
radaronline.com & 0.8805 & 0.1195 & 159 &  & Dave & 0.8869 & 0.1131 & 168 \\
\cline{1-4}\cline{6-9}
\end{tabular}
}
\end{table}

\subsection{System Description}
We have fine-tuned our pre-trained models using AdamW\cite{adamW} optimizer and cross-entropy loss after doing label encoding on the target values. We have applied softmax on the logits produced by each model in order to obtain the prediction probability vectors. The experiments were performed on a system with 16GB RAM and 2.2 GHz Quad-Core Intel Core i7 Processor, along with a Tesla T4 GPU, with batch size of 32. The maximum input sequence length was fixed at 128. Initial learning rate was set to 2e-5. The number of epochs varied from 6 to 15 depending on the model. 

\subsection{Evaluation Metrics}
For evaluation of fake news classification, we have used precision, recall, accuracy and f1-score to measure performance of models. We additionally have used two metric, negative log likelihood (NLL) loss and Brier score for evaluating
predictive uncertainty of the model. More details on these metrics are following.

\textbf{Negative Log Likelihood:} The negative log-likelihood function produces a high value when all the values in a prediction vector are evenly distributed, i.e. when the classification is unclear. It also produces relatively high values in case of wrong classification. However, its value is very small when the output matches the expected value.
\begin{equation}
L_{\log }(y, p)=-(y \log (p)+(1-y) \log (1-p))
\end{equation}
where $p$ is prediction vector and $y$ is the true labels.

\textbf{Brier Score:} The brier score is a metric that is applied for prediction probabilities. It calculates the mean squared error between the predicted probabilities and actual values. It is quite similar in spirit to the log-loss metric, with a major difference being the fact that it is gentler in penalizing inaccurate predictions.
\begin{equation}
B S=\frac{1}{N} \sum_{i=1}^{N}\left(t_{i}-p_{i}\right)^{2}
\end{equation}
where, $t_{i}$ is the predicted probability and $p_{i}$ is the actual outcome.

\subsection{Training Strategy}
We have used XLNet, RoBERTa, XLM-RoBERTa, DeBERTa, ELECTRA and ERNIE 2.0 as backbone models in the case of COVID-19 Fake News dataset, while XLNet, RoBERTa, XLM-RoBERTa, DeBERTa and NewsBERT served as backbone models for the FakeNewsNet dataset. The training procedure is carried out by adding a fully-connected dense layer at the end of each of these pre-trained models and fine-tuning it on the corresponding dataset. For NewsBERT model, we have used BERT with \textit{uncased\_small\_base} pre-trained weights and modified it by stacking few consecutive fully-connected dense layers. We have used higher max sequence length for its embedding layer. We have trained this model with news article text, similarly to other backbone models for longer epochs.
In order to train the Statistical Feature Fusion Network models, we used the feature samples created using prediction vector from ensemble of fine-tuned language models as features as well as the statistical features. The FakeNewsNet dataset is highly imbalanced, with 75\% of the samples be-longing to the ”real” class and 25\% belonging to the ”fake” class. In order to handle this problem of an imbalanced dataset, we have used the KMeans-SMOTE~\cite{last2017oversampling} algorithm, a variation of the Synthetic Minority Oversampling Technique (SMOTE)~\cite{Chawla_2002}. We synthesize new feature samples from the minority class data points, using the feature set obtained using the individual model prediction vectors and statistical features, in order to balance out the imbalance in class distribution without providing any additional information to the model.

\section{Results}

\subsection{Performance of Individual Models}
We have used each fine-tuned model individually to perform “real” vs “fake” classification. Quantitative results for COVID-19 Fake News dataset are tabulated in Table-\ref{single_base_model}. We can see that XLM-RoBERTa, RoBERTa, XLNet and ERNIE 2.0 perform really well on the validation set. However, RoBERTa has been able to produce the best classification results when evaluated on the test set.
We have also evaluated the performance of XLM-RoBERTa, RoBERTa, XLNet, DeBERTa, and NewsBERT on the FakeNewsNet dataset. Corresponding quantitative results are shown in Table \ref{single_base_model2}. NewsBERT has been able to achieve the best results on the validation set, while RoBERTa produces the best results on the test set.

\begin{table}[h]
\centering
\caption{Individual model performance on validation and test set of COVID-19 Fake News Dataset}
\label{single_base_model}
\resizebox{\textwidth}{!}{%
\begin{tabular}{|c|c|c|c|c|c|c|c|c|}
\hline
\multicolumn{1}{|c|}{\multirow{2}{*}{\textbf{Model Name}}} & \multicolumn{4}{c|}{\textbf{Validation Set}} & \multicolumn{4}{c|}{\textbf{Test set}} \\ \cline{2-9} 
\multicolumn{1}{|c|}{} & \multicolumn{1}{c|}{\textbf{Accuracy}} & \multicolumn{1}{c|}{\textbf{Precision}} & \multicolumn{1}{c|}{\textbf{Recall}} & \multicolumn{1}{c|}{\textbf{F1 Score}} & \multicolumn{1}{c|}{\textbf{Accuracy}} & \multicolumn{1}{c|}{\textbf{Precision}} & \multicolumn{1}{c|}{\textbf{Recall}} & \multicolumn{1}{c|}{\textbf{F1 Score}} \\ \hline
XLM-RoBERTa (base) & 0.968 & 0.968 & 0.968 & 0.968 & 0.970 & 0.970 & 0.970 & 0.970 \\ \hline
RoBERTa (base) & 0.970 & 0.970 & 0.970 & 0.970 & \textbf{0.972} & \textbf{0.972} & \textbf{0.972} & \textbf{0.972} \\ \hline
XLNet (base, cased) & 0.975 & 0.975 & 0.975 & 0.975 & 0.966 & 0.966 & 0.966 & 0.966 \\ \hline
DeBERTa (base) & 0.964 & 0.964 & 0.964 & 0.964 & 0.964 & 0.964 & 0.964 & 0.964 \\ \hline
ELECTRA (base) & 0.948 & 0.948 & 0.948 & 0.948 & 0.953 & 0.953 & 0.953 & 0.953 \\ \hline
ERNIE 2.0 & \textbf{0.976} & \textbf{0.976} & \textbf{0.976} & \textbf{0.976} & 0.969 & 0.969 & 0.969 & 0.969 \\ \hline
\end{tabular}%
}
\end{table}

\begin{table}[h]
\centering
\caption{Individual model performance on validation and test set of FakeNewsNet Dataset}
\label{single_base_model2}
\resizebox{\textwidth}{!}{%
\begin{tabular}{|c|c|c|c|c|c|c|c|c|}
\hline
\multicolumn{1}{|c|}{\multirow{2}{*}{\textbf{Model Name}}} & \multicolumn{4}{c|}{\textbf{Validation Set}} & \multicolumn{4}{c|}{\textbf{Test set}} \\ \cline{2-9} 
\multicolumn{1}{|c|}{} & \multicolumn{1}{c|}{\textbf{Accuracy}} & \multicolumn{1}{c|}{\textbf{Precision}} & \multicolumn{1}{c|}{\textbf{Recall}} & \multicolumn{1}{c|}{\textbf{F1 Score}} & \multicolumn{1}{c|}{\textbf{Accuracy}} & \multicolumn{1}{c|}{\textbf{Precision}} & \multicolumn{1}{c|}{\textbf{Recall}} & \multicolumn{1}{c|}{\textbf{F1 Score}} \\ \hline
XLM-RoBERTa (base) & 0.8548 & 0.8548 & 0.8548 & 0.8548 & 0.8631 & 0.8631 & 0.8631 & 0.8631 \\ \hline
RoBERTa (base) & 0.8636 & 0.8636 & 0.8636 & 0.8636 & \textbf{0.8652} & \textbf{0.8652} & \textbf{0.8652} & \textbf{0.8652} \\ \hline
XLNet (base, cased) & 0.8600 & 0.8600 & 0.8600 & 0.8600 & 0.8605 & 0.8605 & 0.8605 & 0.8605 \\ \hline
DeBERTa (base) & 0.8657 & 0.8657 & 0.8657 & 0.8657 & 0.8580 & 0.8580 & 0.8580 & 0.8580 \\ \hline
NewsBERT & \textbf{0.8694} & \textbf{0.8694} & \textbf{0.8694} & \textbf{0.8694} & 0.8626 & 0.8626 & 0.8626 & 0.8626 \\ \hline
\end{tabular}%
}
\end{table}

\subsection{Performance of Ensemble Models}
We tried out different combinations of pre-trained models with both the ensemble techniques: Soft Voting and Hard Voting. 
Performance for different ensembles on the COVID-19 Fake News dataset are shown in Table-\ref{soft_voting_model} and \ref{hard_voting_model}.
From the results, we can infer that the ensemble models significantly outperform the individual models, and Soft-voting ensemble method performed better overall than Hard-voting ensemble method. Hard-voting Ensemble model consisting of RoBERTa, XLM-RoBERTa, XLNet, ERNIE 2.0 and DeBERTa models performed the best among other hard voting ensembles on both validation and test set. Among the Soft Voting Ensembles, the ensemble consisting of RoBERTa, XLM-RoBERTa, XLNet, ERNIE 2.0 and Electra models achieved best accuracy overall on the validation set and 
a combination of XLNet, RoBERTa, XLM-RoBERTa and DeBERTa models produces the best classification result overall on the test set. 

\begin{table}[h]
\centering
\caption{Performance of Soft Voting for different ensemble models on validation and test set of COVID-19 Fake News Dataset}
\label{soft_voting_model}
\resizebox{\textwidth}{!}{%
\begin{tabular}{|c|c|c|c|c|c|c|c|c|} 
\hline
\multirow{2}{*}{\begin{tabular}[c]{@{}c@{}}\textbf{ Ensemble Model}\\\textbf{ Combination} \end{tabular}} & \multicolumn{4}{c|}{\textbf{Validation Set} } & \multicolumn{4}{c|}{\textbf{Test set} } \\ 
\cline{2-9}
 & \textbf{Accuracy}  & \textbf{Precision}  & \textbf{Recall}  & \textbf{F1 Score}  & \textbf{Accuracy}  & \textbf{Precision}  & \textbf{Recall}  & \textbf{F1 Score}  \\ 
\hline
\begin{tabular}[c]{@{}c@{}} RoBERTa+XLM-RoBERTa\\ +XLNet\\ \end{tabular} & 0.9827 & 0.9827 & 0.9827 & 0.9827 & 0.9808 & 0.9808 & 0.9808 & 0.9808 \\ 
\hline
\begin{tabular}[c]{@{}c@{}} RoBERTa+XLM-RoBERTa\\ +XLNet+DeBERT\\ \end{tabular} & 0.9832 & 0.9832 & 0.9832 & 0.9832 & \textbf{0.9831} & \textbf{0.9831} & \textbf{0.9831} & \textbf{0.9831} \\ 
\hline
\begin{tabular}[c]{@{}c@{}} RoBERTa+XLM-RoBERTa\\ +XLNet+ERNIE 2.0\\ +DeBERTa\\ \end{tabular} & 0.9836 & 0.9836 & 0.9836 & 0.9836 & 0.9822 & 0.9822 & 0.9822 & 0.9822 \\ 
\hline
\begin{tabular}[c]{@{}c@{}} RoBERTa+XLM-RoBERTa\\ +XLNet+ERNIE 2.0\\ +Electra\\ \end{tabular} & \textbf{0.9841} & \textbf{0.9841} & \textbf{0.9841} & \textbf{0.9841} & 0.9808 & 0.9808 & 0.9808 & 0.9808 \\
\hline
\end{tabular}
}
\end{table}

\begin{table}[h]
\centering
\caption{Performance of Hard Voting for different ensemble models on validation and test set of COVID-19 Fake News Dataset}
\label{hard_voting_model}
\resizebox{\textwidth}{!}{%
\begin{tabular}{|c|c|c|c|c|c|c|c|c|} 
\hline
\multirow{2}{*}{\begin{tabular}[c]{@{}c@{}}\textbf{ Ensemble Model}\\\textbf{ Combination} \end{tabular}} & \multicolumn{4}{c|}{\textbf{Validation Set} } & \multicolumn{4}{c|}{\textbf{Test set} } \\ 
\cline{2-9}
 & \textbf{Accuracy}  & \textbf{Precision}  & \textbf{Recall}  & \textbf{F1 Score}  & \textbf{Accuracy}  & \textbf{Precision}  & \textbf{Recall}  & \textbf{F1 Score}  \\ 
\hline
\begin{tabular}[c]{@{}c@{}} RoBERTa+XLM-RoBERTa\\ +XLNet\\ \end{tabular} & 0.9818 & 0.9818 & 0.9818 & 0.9818 & 0.9804 & 0.9804 & 0.9804 & 0.9804 \\ 
\hline
\begin{tabular}[c]{@{}c@{}} RoBERTa+XLM-RoBERTa\\ +XLNet+DeBERT\\ \end{tabular} & 0.9748 & 0.9748 & 0.9748 & 0.9748 & 0.9743 & 0.9743 & 0.9743 & 0.9743 \\ 
\hline
\begin{tabular}[c]{@{}c@{}} RoBERTa+XLM-RoBERTa\\ +XLNet+ERNIE 2.0\\ +DeBERTa\\ \end{tabular} & \textbf{0.9832} & \textbf{0.9832} & \textbf{0.9832} & \textbf{0.9832} & \textbf{0.9813} & \textbf{0.9813} & \textbf{0.9813} & \textbf{0.9813} \\ 
\hline
\begin{tabular}[c]{@{}c@{}} RoBERTa+XLM-RoBERTa\\ +XLNet+ERNIE 2.0\\ +Electra\\ \end{tabular} & 0.9822 & 0.9822 & 0.9822 & 0.9822 & 0.9766 & 0.9766 & 0.9766 & 0.9766 \\
\hline
\end{tabular}
}
\end{table}

We have also evaluated our best ensemble model combination from the above approach, consisting of XLM-RoBERTa, RoBERTa, XLNet and DeBERTa, as well as a combination of the above models along with NewsBERT, on the FakeNewsNet dataset in Table \ref{soft_voting_FNN} and \ref{hard_voting_FNN}. We have tried out both soft-voting and hard-voting ensembling techniques, and have observed that the addition of the features obtained from NewsBERT prediction vectors provides a boost to the final F1-score. Also, soft-voting performs slightly better than hard-voting on the test set.

\begin{table}[h]
\centering
\caption{Performance of Soft Voting on validation and test set of FakeNewsNet dataset}
\label{soft_voting_FNN}
\resizebox{\textwidth}{!}{%
\begin{tabular}{|c|c|c|c|c|c|c|c|c|} 
\hline
\multirow{2}{*}{\begin{tabular}[c]{@{}c@{}}\textbf{ Ensemble Model}\\\textbf{ Combination} \end{tabular}} & \multicolumn{4}{c|}{\textbf{Validation Set} } & \multicolumn{4}{c|}{\textbf{Test set} } \\ 
\cline{2-9}
 & \textbf{Accuracy}  & \textbf{Precision}  & \textbf{Recall}  & \textbf{F1 Score}  & \textbf{Accuracy}  & \textbf{Precision}  & \textbf{Recall}  & \textbf{F1 Score}  \\ 
\hline

\begin{tabular}[c]{@{}c@{}} RoBERTa+XLM-RoBERTa\\ +XLNet+DeBERTa\\ \end{tabular} & 0.8699 & 0.8699 & 0.8699 & 0.8699 & 0.8718 & 0.8718 & 0.8718 & 0.8718 \\ 
\hline
\begin{tabular}[c]{@{}c@{}} RoBERTa+XLM-RoBERTa\\ +XLNet+DeBERTa\\ +NewsBERT\\ \end{tabular} & \textbf{0.8783} & \textbf{0.8783} & \textbf{0.8783} & \textbf{0.8783} & \textbf{0.8765} & \textbf{0.8765} & \textbf{0.8765} & \textbf{0.8765} \\
\hline
\end{tabular}
}
\end{table}

\begin{table}[h]
\centering
\caption{Performance of Hard Voting on validation and test set of FakeNewsNet dataset}
\label{hard_voting_FNN}
\resizebox{\textwidth}{!}{%
\begin{tabular}{|c|c|c|c|c|c|c|c|c|} 
\hline
\multirow{2}{*}{\begin{tabular}[c]{@{}c@{}}\textbf{ Ensemble Model}\\\textbf{ Combination} \end{tabular}} & \multicolumn{4}{c|}{\textbf{Validation Set} } & \multicolumn{4}{c|}{\textbf{Test set} } \\ 
\cline{2-9}
 & \textbf{Accuracy}  & \textbf{Precision}  & \textbf{Recall}  & \textbf{F1 Score}  & \textbf{Accuracy}  & \textbf{Precision}  & \textbf{Recall}  & \textbf{F1 Score}  \\ 
\hline
\begin{tabular}[c]{@{}c@{}} RoBERTa+XLM-RoBERTa\\ +XLNet+DeBERT\\ \end{tabular} & 0.8662 & 0.8662 & 0.8662 & 0.8662 & 0.8672 & 0.8672 & 0.8672 & 0.8672 \\ 
\hline
\begin{tabular}[c]{@{}c@{}} RoBERTa+XLM-RoBERTa\\ +XLNet+DeBERT+NewsBERT\\ \end{tabular} & \textbf{0.8783} & \textbf{0.8783} & \textbf{0.8783} & \textbf{0.8783} & \textbf{0.8749} & \textbf{0.8749} & \textbf{0.8749} & \textbf{0.8749} \\ 
\hline
\end{tabular}
}
\end{table}

\subsection{Performance of Statistical Feature Fusion Network and Comparisons}
In this section, we have qualitatively measured the performance of our Statistical Feature Fusion Network (SFFN) with MCDropout with respect to SFFN. We have also compared the performance of various classical models like Logistic Regression, SVM, Decision Tree, Random Forest. As a feature input to SFFN, we have studied two different feature input types. The first type of feature set is created using the individual prediction vector from the various language models of the ensemble (soft-voting), with the conditional probability values of various attributes as statistical features and the second type of feature set is created using the prediction vector from the ensemble (soft-voting) of the language models with the same conditional probability features as the previous one.

In Table \ref{DL_Stat_COVID}, we have experimented with some classical machine learning models on a new feature set created using the individual predictions from the language models of the best ensemble mentioned in Table \ref{soft_voting_model}, and the conditional probability values of URL domains and username handles for the COVID-19 Fake News dataset. In Table \ref{DL_Stat_FNN}, we have tabulated the results of the same experiment, with the best ensemble from the Table \ref{soft_voting_FNN}, on the FakeNewsNet dataset using the conditional probability values of news author and news source.

\begin{table}[h]
\centering
\caption{Individual DL model predictions + Statistical features on validation and test set of COVID-19 Fake News dataset}
\label{DL_Stat_COVID}
\resizebox{\textwidth}{!}{%
\begin{tabular}{|c|c|c|c|c|c|c|c|c|} 
\hline
\multirow{2}{*}{\begin{tabular}[c]{@{}c@{}}\textbf{Model} \end{tabular}} & \multicolumn{4}{c|}{\textbf{Validation Set} } & \multicolumn{4}{c|}{\textbf{Test set} } \\ 
\cline{2-9}
 & \textbf{Accuracy}  & \textbf{Precision}  & \textbf{Recall}  & \textbf{F1 Score}  & \textbf{Accuracy}  & \textbf{Precision}  & \textbf{Recall}  & \textbf{F1 Score}  \\ 
\hline
\begin{tabular}[c]{@{}c@{}} Logistic Regression \end{tabular} & 0.9841 & 0.9841 & 0.9841 & 0.9841 & 0.9832 & 0.9832 & 0.9832 & 0.9832 \\ 
\hline
\begin{tabular}[c]{@{}c@{}} SVM \end{tabular} & 0.9825 & 0.9825 & 0.9825 & 0.9825 & 0.9827 & 0.9827 & 0.9827 & 0.9827 \\ 
\hline
\begin{tabular}[c]{@{}c@{}} Decision Tree \end{tabular} & 0.9804 & 0.9804 & 0.9804 & 0.9804 & 0.9743 & 0.9743 & 0.9743 & 0.9743 \\ 
\hline
\begin{tabular}[c]{@{}c@{}} Random Forest \end{tabular} & 0.9808 & 0.9808 & 0.9808 & 0.9808 & 0.9804 & 0.9804 & 0.9804 & 0.9804 \\
\hline
\begin{tabular}[c]{@{}c@{}} SFFN \end{tabular} & 0.9841 & 0.9841 & 0.9841 & 0.9841 & 0.9822 & 0.9822 & 0.9822 & 0.9822 \\
\hline
\begin{tabular}[c]{@{}c@{}} SFFN with MCDropout \end{tabular} & \textbf{0.9846} & \textbf{0.9846} & \textbf{0.9846} & \textbf{0.9846} & \textbf{0.9836} & \textbf{0.9836} & \textbf{0.9836} & \textbf{0.9836} \\
\hline
\end{tabular}
}
\end{table}

\begin{table}[h]
\centering
\caption{Individual DL model predictions + Statistical features on validation and test set of FakeNewsNet dataset using news title and article text}
\label{DL_Stat_FNN}
\resizebox{\textwidth}{!}{%
\begin{tabular}{|c|c|c|c|c|c|c|c|c|} 
\hline
\multirow{2}{*}{\begin{tabular}[c]{@{}c@{}}\textbf{Model} \end{tabular}} & \multicolumn{4}{c|}{\textbf{Validation Set} } & \multicolumn{4}{c|}{\textbf{Test set} } \\ 
\cline{2-9}
 & \textbf{Accuracy}  & \textbf{Precision}  & \textbf{Recall}  & \textbf{F1 Score}  & \textbf{Accuracy}  & \textbf{Precision}  & \textbf{Recall}  & \textbf{F1 Score}  \\ 
\hline
\begin{tabular}[c]{@{}c@{}} Logistic Regression \end{tabular} & 0.8714 & 0.8714 & 0.8714 & 0.8714 & 0.8644 & 0.8644 & 0.8644 & 0.8644 \\ 
\hline
\begin{tabular}[c]{@{}c@{}} SVM \end{tabular} & 0.8772 & 0.8772 & 0.8772 & 0.8772 & 0.8678 & 0.8678 & 0.8678 & 0.8678 \\ 
\hline
\begin{tabular}[c]{@{}c@{}} Decision Tree \end{tabular} & 0.8647 & 0.8647 & 0.8647 & 0.8647 & 0.8600 & 0.8600 & 0.8600 & 0.8600 \\ 
\hline
\begin{tabular}[c]{@{}c@{}} Random Forest \end{tabular} & 0.8662 & 0.8662 & 0.8662 & 0.8662 & 0.8641 & 0.8641 & 0.8641 & 0.8641 \\
\hline
\begin{tabular}[c]{@{}c@{}} SFFN \end{tabular} & 0.8824 & 0.8824 & 0.8824 & 0.8824 & 0.8678 & 0.8678 & 0.8678 & 0.8678 \\
\hline
\begin{tabular}[c]{@{}c@{}} SFFN with MCDropout \end{tabular} & \textbf{0.8850} & \textbf{0.8850} & \textbf{0.8850} & \textbf{0.8850} & \textbf{0.8728} & \textbf{0.8728} & \textbf{0.8728} & \textbf{0.8728} \\
\hline
\end{tabular}
}
\end{table}

\begin{table}[hbt!]
\centering
\caption{DL soft-voting predictions + Statistical features on validation and test set of COVID-19 Fake News dataset}
\label{DL_soft_Stat_COVID}
\resizebox{\textwidth}{!}{%
\begin{tabular}{|c|c|c|c|c|c|c|c|c|} 
\hline
\multirow{2}{*}{\begin{tabular}[c]{@{}c@{}}\textbf{Model} \end{tabular}} & \multicolumn{4}{c|}{\textbf{Validation Set} } & \multicolumn{4}{c|}{\textbf{Test set} } \\ 
\cline{2-9}
 & \textbf{Accuracy}  & \textbf{Precision}  & \textbf{Recall}  & \textbf{F1 Score}  & \textbf{Accuracy}  & \textbf{Precision}  & \textbf{Recall}  & \textbf{F1 Score}  \\ 
\hline
\begin{tabular}[c]{@{}c@{}} Logistic Regression \end{tabular} & 0.9836 & 0.9836 & 0.9836 & 0.9836 & 0.9831 & 0.9831 & 0.9831 & 0.9831 \\ 
\hline
\begin{tabular}[c]{@{}c@{}} SVM \end{tabular} & 0.9831 & 0.9831 & 0.9831 & 0.9831 & 0.9827 & 0.9827 & 0.9827 & 0.9827 \\ 
\hline
\begin{tabular}[c]{@{}c@{}} Decision Tree \end{tabular} & 0.9832 & 0.9832 & 0.9832 & 0.9832 & 0.9827 & 0.9827 & 0.9827 & 0.9827 \\ 
\hline
\begin{tabular}[c]{@{}c@{}} Random Forest \end{tabular} & 0.9832 & 0.9832 & 0.9832 & 0.9832 & 0.9827 & 0.9827 & 0.9827 & 0.9827 \\
\hline
\begin{tabular}[c]{@{}c@{}} SFFN \end{tabular} & \textbf{0.9841} & \textbf{0.9841} & \textbf{0.9841} & \textbf{0.9841} & 0.9832 & 0.9832 & 0.9832 & 0.9832 \\
\hline
\begin{tabular}[c]{@{}c@{}} SFFN with MCDropout \end{tabular} & \textbf{0.9841} & \textbf{0.9841} & \textbf{0.9841} & \textbf{0.9841} & \textbf{0.9836} & \textbf{0.9836} & \textbf{0.9836} & \textbf{0.9836} \\
\hline
\end{tabular}
}
\end{table}

\begin{table}[hbt!]
\centering
\caption{DL soft-voting predictions + Statistical features on validation and test set of FakeNewsNet dataset using title and document}
\label{SV_Stats_FNN}
\resizebox{\textwidth}{!}{%
\begin{tabular}{|c|c|c|c|c|c|c|c|c|} 
\hline
\multirow{2}{*}{\begin{tabular}[c]{@{}c@{}}\textbf{Model} \end{tabular}} & \multicolumn{4}{c|}{\textbf{Validation Set} } & \multicolumn{4}{c|}{\textbf{Test set} } \\ 
\cline{2-9}
 & \textbf{Accuracy}  & \textbf{Precision}  & \textbf{Recall}  & \textbf{F1 Score}  & \textbf{Accuracy}  & \textbf{Precision}  & \textbf{Recall}  & \textbf{F1 Score}  \\ 
\hline
\begin{tabular}[c]{@{}c@{}} Logistic Regression \end{tabular} & 0.8929 & 0.8929 & 0.8929 & 0.8929 & 0.8914 & 0.8914 & 0.8914 & 0.8914 \\ 
\hline
\begin{tabular}[c]{@{}c@{}} SVM \end{tabular} & 0.8966 & 0.8966 & 0.8966 & 0.8966 & 0.8950 & 0.8950 & 0.8950 & 0.8950 \\ 
\hline
\begin{tabular}[c]{@{}c@{}} Decision Tree \end{tabular} & 0.8939 & 0.8939 & 0.8939 & 0.8939 & 0.8976 & 0.8976 & 0.8976 & 0.8976 \\ 
\hline
\begin{tabular}[c]{@{}c@{}} Random Forest \end{tabular} & 0.8939 & 0.8939 & 0.8939 & 0.8939 & 0.8998 & 0.8998 & 0.8998 & 0.8998 \\
\hline
\begin{tabular}[c]{@{}c@{}} SFFN \end{tabular} & 0.9096 & 0.9096 & 0.9096 & 0.9096 & 0.9094 & 0.9094 & 0.9094 & 0.9094 \\
\hline
\begin{tabular}[c]{@{}c@{}} SFFN with MCDropout \end{tabular} & \textbf{0.9101} & \textbf{0.9101} & \textbf{0.9101} & \textbf{0.9101} & \textbf{0.9115} & \textbf{0.9115} & \textbf{0.9115} & \textbf{0.9115} \\
\hline
\end{tabular}
}
\end{table}

Then, we have evaluated the performance of the same models on another type of feature set for COVID-19 Fake News and FakeNewsNet datasets respectively in Table \ref{DL_soft_Stat_COVID} and \ref{SV_Stats_FNN}.
From these studies, we can conclude that SFFN with MCDropout got better accuracy than the other classical models and the feature set using average prediction vector from best performing ensemble model with statistical feature is more significant than the other one for best performance. 

\textbf{Statistical significance test: } We have performed McNemar's test \cite{mcnemar1947note} between different approaches to check whether the results are statistically significant or not. 
While comparing the individual models to both the Soft-Voting and SFNN approaches, we observe in Tables \ref{mc_1}-\ref{mc_4} that the p-values obtained are always less than the pre-defined significance level. We can thus conclude that the error rates of using these two ensemble approaches are indeed different from using just a single model.

\begin{table}[hbt!]
\centering
\caption{McNemar’s Test Results on FakeNewsNet Dataset with respect to SFFN Model }
\label{mc_1}
\resizebox{\textwidth}{!}{%
\begin{tabular}{|c|c|c|c|c|c|c|} 
\hline
\multirow{2}{*}{\textbf{Model }} & \multicolumn{3}{c|}{\textbf{Validation Set}} & \multicolumn{3}{c|}{\textbf{Test Set}} \\ 
\cline{2-7}
 & \textbf{McNemar’s test statistic } & \textbf{p-value } & \textbf{Reject H0 (alpha=0.05) } & \textbf{McNemar’s test statistic } & \textbf{p-value } & \textbf{Reject H0 (alpha=0.05) } \\ 
\hline
RoBERTa (base) & 59.00 & 0 & True & 55.00 & 0 & True \\ 
\hline
XLM-RoBERTa (base) & 46.00 & 0 & True & 48.00 & 0 & True \\ 
\hline
XLNet (base, cased) & 52.00 & 0 & True & 51.00 & 0 & True \\ 
\hline
DeBERTa (base) & 62.00 & 0 & True & 56.00 & 0 & True \\ 
\hline
NewsBERT & 66.00 & 0 & True & 70.00 & 0 & True \\
\hline
\end{tabular}
}
\end{table}

\begin{table}[hbt!]
\centering
\caption{McNemar’s Test Results on FakeNewsNet Dataset with respect to Soft Voting Ensemble}
\label{mc_2}
\resizebox{\textwidth}{!}{%
\begin{tabular}{|c|c|c|c|c|c|c|} 
\hline
\multirow{2}{*}{\textbf{Model }} & \multicolumn{3}{c|}{\textbf{Validation Set}} & \multicolumn{3}{c|}{\textbf{Test Set}} \\ 
\cline{2-7}
 & \textbf{McNemar’s test statistic } & \textbf{p-value } & \textbf{Reject H0 (alpha=0.05) } & \textbf{McNemar’s test statistic } & \textbf{p-value } & \textbf{Reject H0 (alpha=0.05) } \\ 
\hline
RoBERTa (base) & 37.00 & 0.007 & True & 44.00 & 0.045 & True \\ 
\hline
XLM-RoBERTa (base) & 25.00 & 0 & True & 23.00 & 0.003 & True \\ 
\hline
XLNet (base, cased) & 39.00 & 0.001 & True & 35.00 & 0.003 & True \\ 
\hline
DeBERTa (base) & 29.00 & 0.011 & True & 25.00 & 0 & True \\
\hline
\end{tabular}
}
\end{table}

\begin{table}[hbt!]
\centering
\caption{McNemar’s Test Results on COVID-19 Fake News Dataset with respect to  SFFN Model}
\label{mc_3}
\resizebox{\textwidth}{!}{%
\begin{tabular}{|c|c|c|c|c|c|c|} 
\hline
\multirow{2}{*}{\textbf{Model }} & \multicolumn{3}{c|}{\textbf{Validation Set}} & \multicolumn{3}{c|}{\textbf{Test Set}} \\ 
\cline{2-7}
 & \textbf{McNemar’s test statistic } & \textbf{p-value } & \textbf{Reject H0 (alpha=0.05) } & \textbf{McNemar’s test statistic } & \textbf{p-value } & \textbf{Reject H0 (alpha=0.05) } \\ 
\hline
RoBERTa (base) & 6.00 & 0 & True & 8.00 & 0 & True \\ 
\hline
XLM-RoBERTa (base) & 4.00 & 0 & True & 7.00 & 0 & True \\ 
\hline
XLNet (base, cased) & 10.00 & 0.003 & True & 9.00 & 0 & True \\ 
\hline
DeBERTa (base) & 9.00 & 0 & True & 6.00 & 0 & True \\
\hline
\end{tabular}
}
\end{table}

\begin{table}[hbt!]
\centering
\caption{McNemar’s Test Results on COVID-19 Fake News Dataset with respect to  Soft Voting Ensemble Model}
\label{mc_4}
\resizebox{\textwidth}{!}{%
\begin{tabular}{|c|c|c|c|c|c|c|} 
\hline
\multirow{2}{*}{\textbf{Model }} & \multicolumn{3}{c|}{\textbf{Validation Set}} & \multicolumn{3}{c|}{\textbf{Test Set}} \\ 
\cline{2-7}
 & McNemar’s test statistic & p-value & Reject H0 (alpha=0.05) & McNemar’s test statistic & p-value & Reject H0 (alpha=0.05) \\ 
\hline
RoBERTa (base) & 7.00 & 0 & True & 5.00 & 0 & True \\ 
\hline
XLM-RoBERTa (base) & 4.00 & 0 & True & 12.00 & 0 & True \\ 
\hline
XLNet (base, cased) & 13.00 & 0.014 & True & 11.00 & 0 & True \\ 
\hline
DeBERTa (base) & 6.00 & 0 & True & 5.00 & 0 & True \\
\hline
\end{tabular}
}
\end{table}

\subsection{Comparative Performance of Different Over-sampling Strategies}

We have tried out various techniques for modelling imbalanced datasets, including variations of the vanilla SMOTE~\cite{chawla2002smote} algorithm. ADASYN~\cite{he2008adasyn} focuses on generating samples that are similar to the original samples that were incorrectly classified using a k-Nearest Neighbors classifier, but SMOTE's basic implementation will not distinguish between easy and hard samples to be classified using the nearest neighbours criterion. As a result, the decision function discovered during training will differ between algorithms. Borderline-SMOTE~\cite{han2005borderline} only generates synthetic data along the decision boundary between the two classes unlike SMOTE, which generates synthetic data at random across the two classes. The primary distinction between SVM-SMOTE and other SMOTE is that instead of employing K-nearest neighbours to detect misclassification like in Borderline-SMOTE~\cite{han2005borderline}, SVM-SMOTE~\cite{nguyen2011borderline} would use the SVM algorithm. Before implementing SMOTE, the KMeans-SMOTE~\cite{last2017oversampling} algorithm uses a KMeans clustering method. Depending on the cluster density, clustering will group samples together and generate new samples.
However, as we can observe from Table-\ref{Over-sampling} , the KMeans-SMOTE algorithm has been able to achieve the best results.

\begin{table}[hbt!]
\centering
\caption{Comparative Performance of Different Over-sampling Strategies}
\label{Over-sampling}
\resizebox{\textwidth}{!}{%
\begin{tabular}{|c|c|c|c|c|c|c|c|c|} 
\hline
\multirow{2}{*}{\begin{tabular}[c]{@{}c@{}}\textbf{Over-sampling }\\\textbf{Strategy}\end{tabular}} & \multicolumn{4}{c|}{\textbf{Validation Set}}                                 & \multicolumn{4}{c|}{\textbf{Test set}}                                        \\ 
\cline{2-9}
                                                                                                    & \textbf{Accuracy} & \textbf{Precision} & \textbf{Recall} & \textbf{F1 Score} & \textbf{Accuracy} & \textbf{Precision} & \textbf{Recall} & \textbf{F1 Score}  \\ 
\hline
ADASYN                                                                                              & 0.8934            & 0.8934             & 0.8934          & 0.8934            & 0.9001            & 0.9001             & 0.9001          & 0.9001             \\ 
\hline
Borderline-SMOTE                                                                                    & 0.8924            & 0.8924             & 0.8924          & 0.8924            & 0.9022            & 0.9022             & 0.9022          & 0.9022             \\ 
\hline
SVM~SMOTE                                                                                           & 0.8971            & 0.8971             & 0.8971          & 0.8971            & 0.9007            & 0.9007             & 0.9007          & 0.9007             \\ 
\hline
KMeans-SMOTE                                                                                               & \textbf{0.9101}            & \textbf{0.9101}             & \textbf{0.9101}          & \textbf{0.9101}            & \textbf{0.9115}            & \textbf{0.9115}             & \textbf{0.9115}          & \textbf{0.9115}             \\
\hline
SMOTE                                                                                               & 0.9039            & 0.9039             & 0.9039          & 0.9039            & 0.9053            & 0.9053             & 0.9053          & 0.9053             \\
\hline
\end{tabular}
}
\end{table}

\subsection{Ablation Study on Heuristic Post-processing}
We augmented our Fake News Detection System with an additional heuristic algorithm to boost the accuracy of the model further. We have used the best performing ensemble model consisting of RoBERTa, XLM-RoBERTa, XLNet and DeBERTa for this approach.
We have performed an ablation study by assigning various levels of priority to each of the features (for example, username$>$domain or author$>$source) and then checking which class’s probability value for that feature is maximum for a particular news item, so that we can assign the corresponding “real” or “fake” class label to that particular item. For example, in one iteration, we have given URL domains a higher priority than username handles to select the label class.  Results for different priority and feature set is shown in Table \ref{ablation_tab} and \ref{ablation_tab_FNN}.

Another important parameter that we have introduced for our experiment is a threshold on the class-wise probability values for the features. For example, if the probability that a particular attribute that exists in a news item belongs to “real” class  is greater than that of it belonging to “fake” class, and the probability of it belonging to the “real” class is greater than a specific threshold, we assign a “real” label to the item. The value of this threshold is a hyperparameter that has been tuned based on the classification accuracy on the validation set. We have summarized the results from our study with and without the threshold parameter in Tables \ref{ablation_tab} and \ref{ablation_tab_FNN}.

As we can observe from the results, the URL domain plays a significant role for ensuring a better classification result when the threshold parameter is taken into account in case of COVID-19 Fake news dataset, while the news author plays a significant role in an analogous scenario in case of the FakeNewsNet dataset. The best results are obtained when we consider the threshold parameter and both the username and domain attributes in case of COVID-19 Fake News dataset, and the news author and news source along with the threshold in the case of FakeNewsNet dataset, with higher importances given to the username and news author. 
We have also performed a similar ablation study on the FakeNewsNet dataset using the author and source attributes.

\begin{table}[h]
\centering
\caption{Ablation Study of Heuristic algorithm on COVID-19 Fake News Dataset}
\label{ablation_tab}
\resizebox{\textwidth}{!}{%
\begin{tabular}{|c|c|c|c|c|} 
\hline
\multirow{2}{*}{\begin{tabular}[c]{@{}c@{}}\textbf{Combination of Attributes~}\\(\textbf{in descending order of Attribute Priority}) \end{tabular}} & \multicolumn{2}{c|}{\textbf{with Threshold} } & \multicolumn{2}{c|}{\textbf{without Threshold} } \\ 
\cline{2-5}
 & \begin{tabular}[c]{@{}c@{}}\textbf{F1 Score on}\\\textbf{Validation Set}\end{tabular} & \begin{tabular}[c]{@{}c@{}}\textbf{F1 Score on}\\\textbf{Test Set}\\ \end{tabular} & \begin{tabular}[c]{@{}c@{}}\textbf{F1 Score on}\\\textbf{Validation Set}\\ \end{tabular} & \begin{tabular}[c]{@{}c@{}}\textbf{F1 Score on}\\\textbf{Test Set}\\ \end{tabular} \\ 
\hline
\{\textit{username, ensemble model pred}~\}  & 0.9831 & 0.9836 & \textbf{0.9822} & \textbf{0.9804} \\ 
\hline
\{\textit{domain, ensemble model pred~}\}  & \textbf{0.9917} & 0.9878 & 0.9635 & 0.9523 \\ 
\hline
\{\textit{domain, username, ensemble model pred~}\}  & 0.9911 & 0.9878 & 0.9635 & 0.9519 \\ 
\hline
\{\textit{username, domain, ensemble model pred~}\}  & 0.9906 & \textbf{0.9883} & 0.9645 & 0.9528 \\
\hline
\end{tabular}
}
\end{table}

\begin{table}[h]
\centering
\caption{Ablation Study of Heuristic algorithm on FakeNewsNet Dataset}
\label{ablation_tab_FNN}
\resizebox{\textwidth}{!}{%
\begin{tabular}{|c|c|c|c|c|} 
\hline
\multirow{2}{*}{\begin{tabular}[c]{@{}c@{}}\textbf{Combination of Attributes~}\\(\textbf{in descending order of Attribute Priority}) \end{tabular}} & \multicolumn{2}{c|}{\textbf{with Threshold} } & \multicolumn{2}{c|}{\textbf{without Threshold} } \\ 
\cline{2-5}
 & \begin{tabular}[c]{@{}c@{}}\textbf{F1 Score on}\\\textbf{Validation Set}\end{tabular} & \begin{tabular}[c]{@{}c@{}}\textbf{F1 Score on}\\\textbf{Test Set}\\ \end{tabular} & \begin{tabular}[c]{@{}c@{}}\textbf{F1 Score on}\\\textbf{Validation Set}\\ \end{tabular} & \begin{tabular}[c]{@{}c@{}}\textbf{F1 Score on}\\\textbf{Test Set}\\ \end{tabular} \\ 
\hline
\{\textit{news author, ensemble model pred}~\}  & 0.8824 & 0.8883 & \textbf{0.8626} & \textbf{0.8677} \\ 
\hline
\{\textit{news source, ensemble model pred~}\}  & 0.8887 & 0.8909 & 0.8510 & 0.8533 \\ 
\hline
\{\textit{news source, news author, ensemble model pred~}\}  & \textbf{0.8939} & \textbf{0.9007} & 0.8516 & 0.8543 \\
\hline
\{\textit{news author, news source, ensemble model pred~}\}  & \textbf{0.8939} & 0.9001 & 0.8510 & 0.8595 \\
\hline
\end{tabular}
}
\end{table}

\subsection{Performance of Uncertainty Models}
We evaluate the performance of the predictive uncertainty of Statistical Feature Fusion Network (SFFN) on two mentioned datasets. We have used two proper scoring rules, Brier score and the negative log-likelihood loss where a
lower score corresponds to a better performance in predicting uncertainty value. The scores and respective model accuracy are given in Table \ref{uncertainty_perform} below. The results clearly demonstrate that the SFFN model with Monte Carlo Dropout (MCDropout) leads to both improved predictive uncertainty and accuracy of the respective model for their corresponding datasets.

\begin{table}[h]
\centering
\caption{Performance of Uncertainty Models (SFFN with MC-Dropout)}
\label{uncertainty_perform}
\resizebox{\textwidth}{!}{%
\begin{tabular}{|c|c|c|c|c|c|c|c|c|} 
\hline
\multirow{2}{*}{\textbf{Dataset}}                                           & \multirow{2}{*}{\begin{tabular}[c]{@{}c@{}}\textbf{\textbf{Input}}\\\textbf{\textbf{Feature}}\end{tabular}} & \multirow{2}{*}{\textbf{Model}} & \multicolumn{3}{c|}{\textbf{Validation Set}}                 & \multicolumn{3}{c|}{\textbf{Test set}}                                 \\ 
\cline{4-9}
                                                                            &                                                                                                             &                                 & \textbf{F1 Score} & \textbf{NLL Loss} & \textbf{Brier Score} & \textbf{F1 Score} & \textbf{\textbf{NLL Loss}} & \textbf{Brier Score}  \\ 
\hline
\multirow{4}{*}{\begin{tabular}[c]{@{}c@{}}COVID-19\\FakeNews\end{tabular}} & \multirow{2}{*}{\begin{tabular}[c]{@{}c@{}}Individual DL models\\with Statistical feature\end{tabular}}     & Vanila SFFN                     & 0.9841            & 0.1444            & 0.0158               & 0.9822            & 0.1544                     & 0.0160                \\ 
\cline{3-9}
                                                                            &                                                                                                             & SFFN with MCDropout             & \textbf{0.9846}   & \textbf{0.1792}   & \textbf{0.0144}      & \textbf{0.9836}   & \textbf{0.2250}            & \textbf{0.0156}       \\ 
\cline{2-9}
                                                                            & \multirow{2}{*}{\begin{tabular}[c]{@{}c@{}}DL Ensemble model\\with Statistical feature\end{tabular}}        & Vanila SFFN                     & \textbf{0.9841}   & 0.0910            & 0.0150               & 0.9832            & 0.0949                     & 0.0152                \\ 
\cline{3-9}
                                                                            &                                                                                                             & SFFN with MCDropout             & \textbf{0.9841}   & \textbf{0.0660}   & \textbf{0.0142}      & \textbf{0.9836}   & \textbf{0.0678}            & \textbf{0.0144}       \\ 
\hline
\multirow{4}{*}{FakeNewsNet}                                                & \multirow{2}{*}{\begin{tabular}[c]{@{}c@{}}Individual DL models\\with Statistical feature\end{tabular}}     & Vanila SFFN                     & 0.8813            & 0.5001            & 0.1060               & 0.8718            & 0.5367                     & 0.1134                \\ 
\cline{3-9}
                                                                            &                                                                                                             & SFFN with MCDropout             & \textbf{0.8824}   & \textbf{0.4804}   & \textbf{0.1039}      & \textbf{0.8729}   & \textbf{0.4877}            & \textbf{0.1085}       \\ 
\cline{2-9}
                                                                            & \multirow{2}{*}{\begin{tabular}[c]{@{}c@{}}DL Ensemble model\\with Statistical feature\end{tabular}}        & Vanila SFFN                     & 0.9096            & 0.3124            & 0.0739               & 0.9094   & 0.2997                     & 0.0731                \\ 
\cline{3-9}
                                                                            &                                                                                                             & SFFN with MCDropout             & \textbf{0.9101}   & \textbf{0.2679}   & \textbf{0.0727}      & \textbf{0.9115}   & \textbf{0.2459}            & \textbf{0.0697}       \\
\hline
\end{tabular}
}
\end{table}




\subsection{Performance of Final Proposed Model and Comparisons}
We qualitatively evaluate the performance of the proposed method on the two mentioned datasets. In Table-\ref{Ensemble_SFFN_post_proc}, we have shown that with the addition of feature fusion network, the performance of the framework has improved compared to other models and achieved state of the art results on both datasets. In our earlier work~\cite{das2021heuristic}, we had shown that the heuristic post-processing approach improves the classification accuracy on the test set significantly. However, the incorporation of uncertainty estimation improves the model performance even more.

\begin{table}[h]
\centering
\caption{Performance of Ensemble SFFN (with MC-Dropout) Model combined with Heuristic Post-Processing on Different Datasets}
\label{Ensemble_SFFN_post_proc}
\resizebox{\textwidth}{!}{%
\begin{tabular}{|c|c|c|c|c|} 
\hline
\textbf{Dataset} & \textbf{Model} & \begin{tabular}[c]{@{}c@{}}\textbf{Combination of Attributes\textasciitilde{}}\\\textbf{(in descending order of Attribute Priority) }\end{tabular} & \begin{tabular}[c]{@{}c@{}}\textbf{F1 Score on}\\\textbf{Validation Set }\end{tabular} & \begin{tabular}[c]{@{}c@{}}\textbf{F1 Score on}\\\textbf{Test Set}\\\end{tabular} \\ 
\hline
\multirow{2}{*}{\begin{tabular}[c]{@{}c@{}}COVID-19\\FakeNews\end{tabular}} & Soft Voted~Ensemble Model & \{~\textit{username, domain, ensemble model pred~}\} & 0.9906 & 0.9883 \\ 
\cline{2-5}
 & Ensemble SFFN (with MCDropout) & \{~\textit{username, domain, ensemble\_SFFN\_mcdropout pred~}\} & \textbf{0.9911} & \textbf{0.9892} \\ 
\hline
\multirow{2}{*}{FakeNewsNet} & Soft Voted~Ensemble Model & \{~\textit{news source, news author, ensemble model pred~}\} & 0.8939 & 0.9007 \\ 
\cline{2-5}
 & Ensemble SFFN (with MCDropout) & \{~\textit{news source, news author, ensemble\_SFFN\_mcdropout~pred~}\} & \textbf{0.9112} & \textbf{0.9156} \\
\hline
\end{tabular}
}
\end{table}

We have shown the comparison of the results on the test set obtained by our model before and after applying the post-processing technique against the top 3 teams in the leaderboard for the COVID-19 Fake News dataset in Table \ref{post proc Comp1}. We have also compared the same with state-of the-art approaches on FakeNewsNet Dataset in Table \ref{post proc Comp2}. 

\begin{table}[!h]
\small
\centering
\caption{Performance comparison on test set for COVID-19 Fake News Dataset}
\label{post proc Comp1}
\begin{tabular}{|c|c|c|c|c|} 
\hline
 \textbf{Method}  & \textbf{Accuracy}  & \textbf{Precision}  & \textbf{Recall}  & \textbf{F1 Score}  \\ 
\hline
Team g2tmn (\textit{Rank 1}~\cite{patwa2021overview, glazkova2020g2tmn}) & 0.9869 & 0.9869 & 0.9869 & 0.9869 \\ 
\hline
Team saradhix (\textit{Rank 2}~\cite{patwa2021overview}) & 0.9864 & 0.9865 & 0.9864 & 0.9864 \\ 
\hline
Team xiangyangli (\textit{Rank 3}~\cite{patwa2021overview}) & 0.9860 & 0.9860 & 0.9860 & 0.9860 \\ 
\hline
\begin{tabular}[c]{@{}c@{}}Ensemble Model + Heuristic\\Post-Processing (Das et al.~\cite{das2021heuristic}) \end{tabular} & 0.9883 & 0.9883 & 0.9883 & 0.9883 \\
\hline
\begin{tabular}[c]{@{}c@{}}SFFN (with MCDropout) +\\Heuristic Post-Processing\end{tabular} & \textbf{0.9892} & \textbf{0.9892} & \textbf{0.9892} & \textbf{0.9892} \\
\hline
\end{tabular}
\end{table}

\begin{table}[!h]
\small
\centering
\caption{Performance comparison on test set for FakeNewsNet Dataset}
\label{post proc Comp2}
\begin{tabular}{|c|c|c|c|c|} 
\hline
 \textbf{Method}  & \textbf{Accuracy}  & \textbf{Precision}  & \textbf{Recall}  & \textbf{F1 Score}  \\ 
\hline
FakeFlow ~\cite{ghanem2021fakeflow}) & 0.82 & 0.82 & 0.82 & 0.82 \\ 
\hline
One-Hot LR~\cite{shu2020fakenewsnet}) & 0.7670 & 0.7670 & 0.7670 & 0.7670 \\ 
\hline
FakeNewsTracker~\cite{shu2019fakenewstracker}) & 0.7186 & 0.7186 & 0.7186 & 0.7186 \\ 
\hline
\begin{tabular}[c]{@{}c@{}}Ensemble Model + Heuristic\\Post-Processing \end{tabular} & 0.9007 & 0.9007 & 0.9007 & 0.9007 \\
\hline
\begin{tabular}[c]{@{}c@{}}SFFN (with MCDropout) +\\Heuristic Post-Processing\end{tabular} & \textbf{0.9156} & \textbf{0.9156} & \textbf{0.9156} & \textbf{0.9156} \\
\hline
\end{tabular}
\end{table}

\subsection{Qualitative Study}
Table \ref{result_tab} shows a two examples where the post-processing algorithm corrects the initial prediction in the case of COVID-19 Fake News dataset. The first example is corrected due to extracted domain which is \textit{``news.sky"} and the second one is corrected because of presence of the username handle, \textit{``@drsanjaygupta"}. 
The last two examples stand incorrect even after application of the post-processing algorithm. The first of these can be attributed to the lack of any username handle or URL domain in the text, and also to the fact that DNA Vaccine is a scientific topic that requires some amount of domain knowledge that the model is unable to capture. The second unsuccessful example might be due to the fact that the URL domain present in the text is not a good indicator of its genuineness.

\begin{table}[!t]
\caption{Qualitative comparison between our SFNN(initial) and post-processing(final) approach for COVID-19 Fake News Dataset.}
\centering
\label{result_tab}
\resizebox{\textwidth}{!}{%
\footnotesize
\begin{tabular}{lccc}
\multicolumn{1}{c}{\uline{\textbf{Tweet}}} & \begin{tabular}[c]{@{}c@{}}\textbf{\uline{Initial }}\\\textbf{\uline{Classification }}\\\textbf{\uline{Output}}\end{tabular} & \begin{tabular}[c]{@{}c@{}}\textbf{\uline{Final }}\\\textbf{\uline{Classification }}\\\textbf{\uline{Output}}\end{tabular} & \uline{\textbf{Ground Truth}}  \\ 
\hline
\begin{tabular}[c]{@{}l@{}}Coronavirus: Donald Trump ignores COVID-19 rules \\with 'reckless and selfish' indoor rally https://t.co/JsiHGLMwfO\end{tabular} & fake & real & real \\ 
\hline
\begin{tabular}[c]{@{}l@{}}We're LIVE talking about COVID-19 (a vaccine transmission)\\ with @drsanjaygupta. Join us and ask some questions\\ of your own: https://t.co/e16G2RGdkA https://t.co/Js7lemT1Z6\end{tabular} & real & fake & fake \\ 
\hline
\begin{tabular}[c]{@{}l@{}}*DNA Vaccine: injecting genetic material into the host so that \\host cells create proteins that are similar to those in the virus \\against which the host then creates antibodies\end{tabular} & fake & fake & real \\ 
\hline
\begin{tabular}[c]{@{}l@{}}Early action and social trust are among the reasons \\for Vermont’s low numbers of coronavirus cases. \\https://t.co/lQzAsc6gSG\end{tabular} & real & real & fake
\end{tabular}
}
\end{table}

\begin{table}[!t]
\centering
\caption{Qualitative comparison between our SFNN (initial) and post-processing (final) approach for FakeNewsNet Dataset.}
\label{result_tab_fakenewsnet}
\resizebox{\textwidth}{!}{%
\begin{tabular}{lccc}
\multicolumn{1}{c}{\uline{\textbf{Tweet}}}                                                                                                               & \begin{tabular}[c]{@{}c@{}}\textbf{\uline{Initial }}\\\textbf{\uline{Classification }}\\\textbf{\uline{Output}}\end{tabular} & \begin{tabular}[c]{@{}c@{}}\textbf{\uline{Final }}\\\textbf{\uline{Classification }}\\\textbf{\uline{Output}}\end{tabular} & \uline{\textbf{Ground Truth}}  \\ 
\hline
\begin{tabular}[c]{@{}l@{}}Kendall Jenner’s BF Blake Griffin Labeled A ‘Terrible Kisser’ \\By Olympian Lolo Jones\end{tabular}                           & fake                                                                                                                         & real                                                                                                                       & real                           \\ 
\hline
\begin{tabular}[c]{@{}l@{}}Tristan Thompson Fined \$25K, Not Suspended for Altercation \\at End of Game 1\end{tabular}                                   & real                                                                                                                         & fake                                                                                                                       & fake                           \\ 
\hline
\begin{tabular}[c]{@{}l@{}}Megyn Kelly Doesn't Understand Why People Thought LuAnn \\De Lesseps' Diana Ross Costume Was Considered 'Racist'\end{tabular} & fake                                                                                                                         & fake                                                                                                                       & real                           \\ 
\hline
Gigi Hadid pulls out of Victoria's Secret Fashion Show                                                                                                   & real                                                                                                                         & real                                                                                                                       & fake                          
\end{tabular}
}
\end{table}

In the case of FakeNewsNet Dateset, we have shown the qualitative results in Table~\ref{result_tab_fakenewsnet}. The first two examples indicate how the presence of authors in the text help to correct the initial model predictions. In the last two examples, however, the final output stands incorrect even after application of the post-processing algorithm. It is mainly due to the fact that the frequency of these particular authors and sources in the overall dataset is very low, hence the statistical information conveyed by them regarding the genuineness of the news items are unreliable.

\section{Conclusion}
In this paper, we have proposed a robust framework for identification of fake news items, which can go a long way in eliminating the spread of misinformation on sensitive topics. In our initial approach, we have tried out various pre-trained language models. Our results have significantly improved when we implemented an ensemble mechanism with Soft-voting by using the prediction vectors from various combinations of these models. Furthermore, we have been able to augment our system with a statistical feature fusion network and a novel heuristics-based post-processing algorithm by incorporation of statistical features, that has drastically improved the fake tweet detection accuracy. Our novel heuristic approach shows that meta-attributes like username handle, URL domain, news author, news source, etc. form very important features of news and analyzing them accurately can go a long way in creating a robust framework for fake news detection. We have also quantified the model uncertainty in the task of fake news detection by applying Monte Carlo dropout as a Bayesian approximation in the statistical feature fusion network. With empirical experiments, we have shown the overall performance increase after including uncertainty in the model.

Finally, we would like to pursue more research into how to extend our framework for an active learning based approach by utilizing uncertainty values, and incorporate other combinations of meta-attributes in our model perform on the given datasets. It would be really interesting to evaluate how our system performs on other generic Fake News datasets and also if different values of the threshold parameter for our post-processing system would impact its overall performance.

%

\bibliography{ref}

\end{document}